\documentclass{bmvc2k}

\usepackage{booktabs}   % \toprule, \midrule, \bottomrule, \cmidrule
\usepackage{amsmath}     % math environments
\usepackage{amssymb}     % \checkmark
\usepackage{graphicx}    % \includegraphics
\usepackage{xcolor}      % colours in text/tables
\usepackage{wrapfig}
\title{Addressing Detail Bottlenecks in Latent Diffusion for RGB-to-SWIR Image Translation
}

\addauthor{Kaili Wang}{}{1}
\addauthor{Martin Dimitrievski}{}{2}
\addauthor{Jose Maria Salvador}{}{3}
\addauthor{Ben Stoffelen}{}{1}
\addauthor{David Van Hamme}{}{2}
\addauthor{Lore Goetschalckx}{}{1}

\addinstitution{
imec, 3001 Leuven, Belgium
}
\addinstitution{
imec-IPI-Ghent University, 9000 Ghent, Belgium
}
\addinstitution{
Yale University, New Haven, CT, USA
}

\runninghead{Wang et al.}{Detail Bottlenecks in LDMs for RGB-to-SWIR Translation}

\def\etal{\emph{et al}\bmvaOneDot}

\begin{document}

\maketitle

\begin{figure*}[t]
\centering
\includegraphics[width=0.95\linewidth]{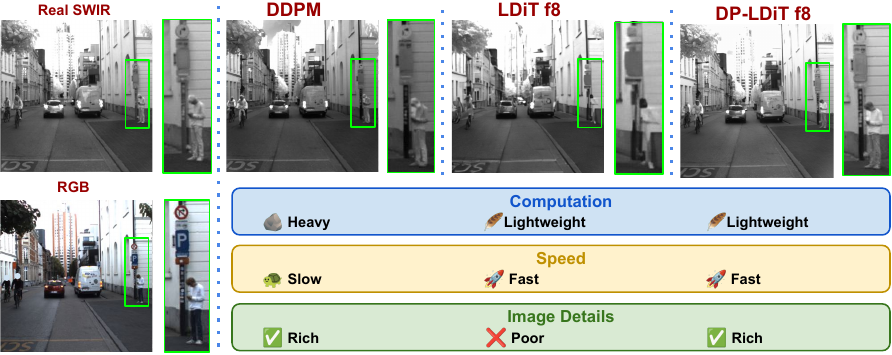}
\caption{Standard latent diffusion (LDiT~$f$8) loses fine details
such as pedestrians and traffic signs (green insets), while our
DP-LDiT~$f$8 recovers them, closely matching real SWIR and the
heavier pixel-space DDPM\@.  Bottom-right table: DP-LDiT
combines the efficiency of latent diffusion with the detail
preservation of pixel-space models.}
\label{fig:teaser}
\vspace{-4mm}
\end{figure*}

\begin{abstract}
Latent diffusion models (LDMs) enable efficient image-to-image
translation but discard fine spatial details during compression,
degrading downstream perception tasks.  We identify two bottlenecks:
the \emph{autoencoder}, which loses spatial information, and the
\emph{conditioning pathway}, which further degrades the source
signal through naive downsampling.  We propose two lightweight,
backbone-agnostic fixes: a \emph{Source-Conditioned Autoencoder}
(SCAE) that injects high-resolution source features into the
decoder via skip connections, and a \emph{Learnable Guidance
Encoder} (LGE) that replaces naive downsampling with a learned
conditioning signal.  Evaluated on RGB-to-SWIR translation
for driving scenes with two denoiser backbones (U-Net and DiT),
our approach improves detection mAP by
up to $2\times$ over the latent diffusion baseline, with up to
$3.4\times$ gains on small objects (COCO-small, ${<}32^2$\,px$^2$),
while achieving state-of-the-art
FID\@.  We further show that FID and detection performance are
poorly correlated, motivating multi-axis evaluation.  Results
generalise zero-shot to the public RASMD
benchmark~\cite{jin2025rasmd}.  We will publicly release
test data with annotations, all checkpoints, and training code.
\end{abstract}

%--------------------------------------------------------------
\section{Introduction}
\label{sec:intro}
%--------------------------------------------------------------

Robust vision systems depend on both the sensing modality and the
perception pipeline built on top of it.  While RGB cameras are
ubiquitous, they degrade under adverse conditions such as low
light, glare, and fog.  This has motivated growing interest in
alternative sensing modalities, in particular Short-Wave Infrared
(SWIR), that capture complementary physical phenomena and can
improve perception in safety-critical
applications~\cite{pinchon2018allweather,jin2025rasmd}.
Understanding how these modalities interact with downstream
perception modules is essential, yet progress is hampered by
a fundamental data bottleneck: collecting and annotating
large-scale datasets for new sensors is expensive and
slow~\cite{goetschalckx2026genai}.

Image-to-image (I2I) translation~\cite{isola2017pix2pix,zhu2017cycle}
offers a practical mechanism to bridge this gap by mapping
abundant RGB data to a target
modality~\cite{lee2023edge,wang2024thermal,jin2025rasmd}.
Diffusion models~\cite{ho2020ddpm,rombach2022ldm} are the
current state of the art, but a fundamental tension exists
between fidelity and efficiency.
Pixel-space models such as DDPM~\cite{ho2020ddpm} preserve spatial
detail but are memory-intensive: at $512 \times 512$ with source
conditioning, a batch size of only~1 fits on an 80\,GB GPU.
Latent diffusion models
(LDMs)~\cite{rombach2022ldm} solve the efficiency problem by
compressing images into a lower-dimensional space before
diffusion, but the compression step loses fine-grained
structure: small objects become amorphous and distant
pedestrians dissolve into blobs.

This loss has direct consequences for perception.  When
translated images are used to evaluate perception systems,
corrupted details degrade downstream task
performance.  Yet the standard generative metric,
Fr\'{e}chet Inception Distance (FID)~\cite{heusel2017fid}, is
poorly suited to detect this problem: it captures global
distributional similarity insensitive to fine local structure
and is computed using features from an RGB-pretrained
network, making its relevance to cross-modal translation
questionable~\cite{jayasumana2024fid}.  We therefore
complement FID with a downstream detection protocol that
directly measures whether translated images preserve the
structural details needed for perception
(Section~\ref{sec:detection}).

We trace the structural fidelity problem in LDMs to two distinct
bottlenecks and propose a targeted fix for each:
\begin{itemize}
\item \textbf{Autoencoder bottleneck.}  The first-stage decoder
cannot recover fine details lost during compression.  Our
\textbf{Source-Conditioned Autoencoder (SCAE)} adds a mirror encoder
$E^{\text{src}}$ whose multi-scale features are injected into the
decoder $D$ via skip connections, giving it direct access to
high-resolution source information at decode time.
\item \textbf{Conditioning bottleneck.}  Naively downsampling the
source image to the latent resolution discards task-relevant
structure.  Our \textbf{Learnable Guidance Encoder (LGE)},
$\tau_\psi$, replaces this with a lightweight convolutional encoder
co-optimised with the denoiser, producing a conditioning signal
concatenated channel-wise with the denoiser input.
\end{itemize}
Both modifications are \textbf{agnostic to the denoiser backbone}.
We demonstrate them with a U-Net denoiser~\cite{ronneberger2015unet}
(\textbf{Detail-Preserving LDM, DP-LDM}) and a
Diffusion Transformer~\cite{peebles2023dit}
(\textbf{Detail-Preserving LDiT, DP-LDiT}),
obtaining consistent improvements with minimal overhead
(+0.15\,GB for DP-LDiT, +0.19\,GB for DP-LDM).
We evaluate on RGB-to-SWIR translation using our own paired
driving dataset and show that our approach improves detection
mAP by up to $2\times$ over latent diffusion baselines, with
up to $3.4\times$ gains on small objects, while simultaneously
achieving state-of-the-art FID\@.  Results transfer zero-shot
to the public RASMD benchmark~\cite{jin2025rasmd}.

Our contributions:
\begin{enumerate}
\item A diagnosis of two distinct detail bottlenecks in latent
diffusion I2I translation (autoencoder compression and conditioning
pathway), together with two backbone-agnostic architectural
modifications (SCAE and LGE) that address them.
\item A downstream detection evaluation protocol that exposes
failure modes invisible to FID, including the finding that FID and
detection performance are poorly correlated, motivating
multi-axis evaluation of cross-modal I2I translation.
\item Comprehensive experiments on RGB-to-SWIR translation showing
large improvements in both distributional and structural fidelity
across two denoiser backbones, with zero-shot transfer to the
public RASMD benchmark~\cite{jin2025rasmd}.
\end{enumerate}

%--------------------------------------------------------------
\section{Related Work}
\label{sec:related}
%--------------------------------------------------------------

\paragraph{Image-to-image translation.}
GAN-based methods such as Pix2pix~\cite{isola2017pix2pix},
Pix2pixHD~\cite{wang2018pix2pixhd}, and
CycleGAN~\cite{zhu2017cycle} established the
paradigm for paired and unpaired I2I translation.  Later advances
such as MUNIT~\cite{huang2018munit} improved diversity through
multimodal mappings.  Diffusion models~\cite{dhariwal2021diffbeat}
subsequently achieved higher sample quality:
Palette~\cite{saharia2022palette} conditions on the source image;
BBDM~\cite{li2023bbdm} leverages Brownian bridge processes.
However, pixel-space diffusion remains computationally expensive,
and none of these works explicitly address the loss of fine spatial
detail that arises when moving to latent-space methods, a gap our
work targets directly.

\paragraph{Latent diffusion and detail preservation.}
LDMs~\cite{rombach2022ldm} reduce cost by encoding images
into a compact latent space with a VQGAN~\cite{esser2021taming}
before running diffusion.  While effective for text-to-image
synthesis, the lossy compression hampers structural fidelity in
I2I tasks~\cite{parmar2024i2iturbo,berrada2025lpl,zhang2025diffusion4k}.
Several recent works attempt to mitigate this.
Berrada~\etal~\cite{berrada2025lpl} propose a latent perceptual
loss (LPL) to encourage sharpness, but applying perceptual losses
during diffusion training requires unreliable one-step clean-image
approximations, yielding limited improvements.
I2I-turbo~\cite{parmar2024i2iturbo} adds encoder--decoder skip
connections in a one-step distilled model, achieving strong
structural preservation when the target is RGB\@.  However, it
relies on a frozen SD-Turbo backbone~\cite{sauer2024sdturbo}
pretrained exclusively on RGB data; when the target modality
differs from RGB, this backbone cannot capture the new domain's
appearance, leading to outputs that retain RGB-like characteristics.
ControlNet~\cite{zhang2023controlnet} adds spatial conditioning to
a frozen text-to-image model via skip connections into the
\emph{denoiser}, but similarly requires a pretrained backbone.
Our approach differs along two axes: (i)~the SCAE skip connections
feed into the first-stage \emph{decoder} rather than the denoiser,
addressing the autoencoder bottleneck that these methods leave
untouched; (ii)~all components are trained from scratch on the
target modality, making the method compatible with any denoiser
backbone and any target domain.

\paragraph{RGB to non-RGB translation.}
Cross-modality translation has been explored for
thermal~\cite{kniaz2019thermalgan,wang2024thermal,lee2023edge},
NIR~\cite{aslahishahri2021nir,jin2025pix2next},
infrared~\cite{ozkanoglu2022infragan}, and
SWIR~\cite{jin2025rasmd} targets.
Jin~\etal~\cite{jin2025rasmd} introduced the RASMD benchmark for
RGB--SWIR translation and established baseline results for
standard models.  However, existing cross-modal
methods adopt off-the-shelf architectures without addressing the
detail bottleneck inherent to latent diffusion, and evaluate
primarily with perceptual metrics (FID, LPIPS) rather than
downstream task performance.  We address both gaps: our
architectural modifications specifically target detail
preservation in latent-space translation, and our evaluation
protocol measures the impact on downstream object detection.

%--------------------------------------------------------------
\section{Method}
\label{sec:method}
%--------------------------------------------------------------

\subsection{Preliminaries}
We frame RGB-to-SWIR translation as learning the conditional
distribution $p(I^{\text{tgt}} \mid I^{\text{src}})$, where
$I^{\text{src}}$ is an image from the source modality (RGB) and
$I^{\text{tgt}}$ from the target modality (SWIR).
A denoising model $\epsilon_\theta$ is trained with the score
matching objective~\cite{song2021sde}:
\begin{equation}
\mathcal{L}_{\text{diff}} =
\mathbb{E}_{I^{\text{tgt}},\,\epsilon\sim\mathcal{N}(0,\mathbf{I})}
\left[\,
\left\|\epsilon - \epsilon_\theta\!\left(z^{\text{tgt}}_t,\, t,\, z^{\text{src}}\right)\right\|_2^2
\,\right],
\label{eq:diff}
\end{equation}
where $z^{\text{tgt}}_t$ is the noisy latent code at timestep $t$ and
$z^{\text{src}}$ is a latent-resolution representation of the source
image.  In standard LDMs, this is obtained by naively resizing
$I^{\text{src}}$; our LGE (Section~\ref{sec:lge}) replaces this with
a learned encoding.  Following standard LDM practice~\cite{rombach2022ldm}, a
frozen VQGAN encoder $E$ maps target images to latents
$z^{\text{tgt}} = E(I^{\text{tgt}})$, and the decoder $D$
reconstructs full-resolution outputs.  Training proceeds in two
stages: first the autoencoder, then the denoiser.

\subsection{Architecture Overview}
\label{sec:bottleneck_method}

\begin{figure*}[t]
\centering
% Replace with actual architecture diagram.
\includegraphics[width=\linewidth]{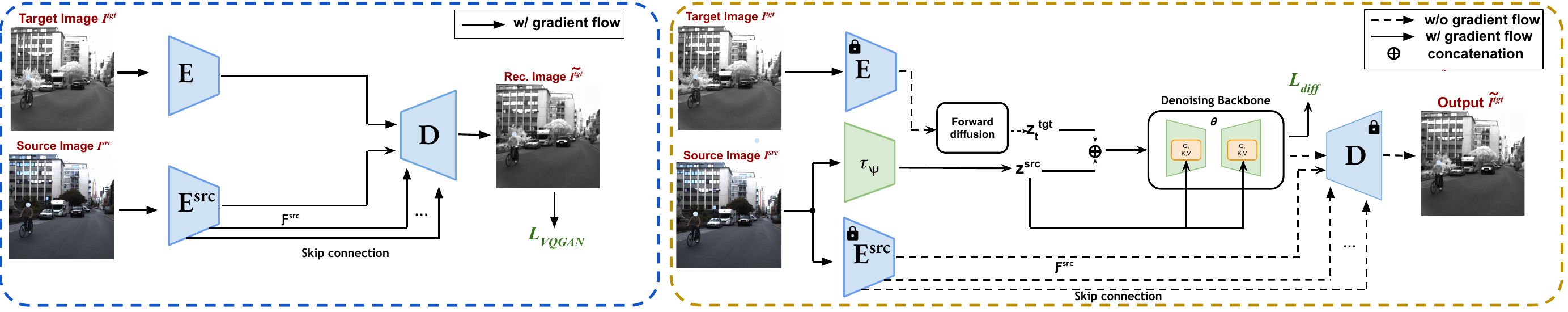}
\vspace{-5mm}
\caption{Overview of the DP-LDM/DP-LDiT pipeline.  The LGE
($\tau_\psi$) encodes the source image into a latent-resolution
conditioning signal for the denoiser.  The SCAE provides skip
connections from $E^{\text{src}}$ to the decoder $D$, injecting
high-resolution source features at decode time.  Frozen modules
are marked with a lock icon.  The denoiser backbone (grey box) can
be either a U-Net (DP-LDM) or a DiT (DP-LDiT).}
\label{fig:architecture}
\end{figure*}

Figure~\ref{fig:architecture} illustrates the complete pipeline.
We identify two distinct sources of detail loss in the standard
LDM pipeline for I2I translation and address each with a targeted
module.  First, the VQGAN autoencoder discards spatial information
during compression; even encoding and immediately decoding a real
image (no diffusion) visibly degrades fine structures such as
distant pedestrians and sign text (quantified in
Section~\ref{sec:bottleneck_exp}).  Second, the conditioning
pathway naively resizes the source image to the latent resolution,
discarding task-relevant structure before the denoiser sees it.
Our \textbf{Source-Conditioned Autoencoder (SCAE)} addresses the
first bottleneck at the decoder level (first training stage),
while our \textbf{Learnable Guidance Encoder (LGE)} addresses the
second at the denoiser level (second training stage).  Both
modifications are \emph{independent of the denoiser backbone}.
We instantiate two variants:
\textbf{DP-LDM} with a U-Net denoiser~\cite{ronneberger2015unet}
and \textbf{DP-LDiT} with a Diffusion
Transformer~\cite{peebles2023dit}.
We use \textbf{LDiT} to denote a standard latent diffusion model
with a DiT backbone (analogous to LDM with a U-Net); both share
the same first-stage autoencoder and latent space, differing only
in the denoiser architecture.
Full architecture specifications are provided in the supplementary
material.

\subsection{Source-Conditioned Autoencoder (SCAE)}
\label{sec:scae}
The VQGAN decoder $D$ reconstructs the full-resolution output
from latent codes, but compression limits its ability to recover
fine details that were present in the high-resolution source.
We add a source encoder $E^{\text{src}}$ that mirrors $D$'s
architecture and extracts multi-scale features
$\mathcal{F}^{\text{src}} = \{F_1^{\text{src}}, \ldots, F_L^{\text{src}}\}$
from $I^{\text{src}}$.  These are injected into $D$ via skip
connections at matching spatial resolutions:
\begin{equation}
\tilde{I}^{\text{tgt}} = D\!\left(E(I^{\text{tgt}}),\;
\mathcal{F}^{\text{src}}\right).
\end{equation}
The input encoder $E$, the source encoder $E^{\text{src}}$, and
the decoder $D$ together form the SCAE\@.
It is trained end-to-end on the target domain following the
VQGAN framework~\cite{esser2021taming}:
\begin{equation}
\mathcal{L}_{\text{VQGAN}}(I^{\text{tgt}},\,\tilde{I}^{\text{tgt}}),
\end{equation}
where
$\tilde{I}^{\text{tgt}} = D(E(I^{\text{tgt}}),\, E^{\text{src}}(I^{\text{src}}))$.
The loss comprises reconstruction, codebook, and adversarial
regularisation terms.
By minimising $\mathcal{L}_{\text{VQGAN}}$, the SCAE learns to
suppress irrelevant source information while selectively
extracting cues from $I^{\text{src}}$ that aid in
reconstructing~$I^{\text{tgt}}$.
The SCAE is frozen when training the denoiser (second stage).

\subsection{Learnable Guidance Encoder (LGE)}
\label{sec:lge}
To address the conditioning bottleneck described in
Section~\ref{sec:bottleneck_method}, we replace the naive source
resizing with a learnable guidance encoder $\tau_\psi$:
\begin{equation}
z^{\text{src}} = \tau_\psi(I^{\text{src}}),\quad
I^{\text{src}} \in \mathbb{R}^{3 \times 512 \times 512},\;
z^{\text{src}} \in \mathbb{R}^{C \times H \times W}.
\end{equation}
$\tau_\psi$ comprises several convolutional and instance-normalisation
layers.  Its parameters $\psi$ are co-optimised with $\theta$ using
$\mathcal{L}_{\text{diff}}$.  The conditioning signal $z^{\text{src}}$
is concatenated channel-wise with the noisy latent $z^{\text{tgt}}_t$
as input to the denoiser.

\vspace{-5mm}
\paragraph{Inference.}
At inference, the source image $I^{\text{src}}$ is processed by
$E^{\text{src}}$ and $\tau_\psi$ to extract features
$\mathcal{F}^{\text{src}}$ and $z^{\text{src}}$, respectively.
Random Gaussian noise is sampled to form $z^{\text{tgt}}_{t=T}$.
At each denoising step, $z^{\text{src}}$ is concatenated
channel-wise with the current noisy latent and passed to the
denoiser.  After $T$ steps, the resulting clean latent is
decoded by $D$ together with $\mathcal{F}^{\text{src}}$ to produce
the final output image $\tilde{I}^{\text{tgt}}$.

% --- Architecture diagram (suggested) ---
%%%%%%%%%%%%%%%%%%%%%%%%%%%%%%%%%%%%%%%%%%%%%%%%%%
%%%%%%%%%%%%%%%%%%%%%%%%%%%%%%%%%%%%%%%%%%%%%%%%%%
%--------------------------------------------------------------
\section{Experiments}
\label{sec:experiments}
%--------------------------------------------------------------

\subsection{Dataset and Evaluation}
\label{sec:dataset}

\paragraph{Dataset.}
We use our own paired RGB--SWIR driving dataset captured in a
medium-sized European city.  RGB and narrow-band SWIR
images were recorded simultaneously using synchronised,
vehicle-mounted cameras.  Image pairs are spatially registered via
homography estimation.  The dataset contains \textbf{31,999
training pairs} recorded in May (Day~1) and two held-out test
splits: \textbf{900 pairs} from Day~1 and \textbf{1,799 pairs}
recorded in June in a different part of the city (Day~2), under
similar weather conditions.  Training and Day~1 test routes do
not overlap; Day~2 covers an entirely different neighbourhood,
providing a geographically and temporally disjoint evaluation.
All images have a native resolution
of $1024 \times 1280$.  During training, images are resized
(preserving aspect ratio) to a random height for scale
augmentation, then randomly cropped to $512 \times 512$ with
random horizontal flipping.  At test time, images are resized
to a height of 512 pixels (preserving aspect ratio) and then
center-cropped to $512 \times 512$.

Both test splits come with professionally produced bounding-box
annotations for three object classes: \emph{Pedestrian},
\emph{Vehicle}, and \emph{Cyclist}.
The Day~1 test set contains \textbf{6,906} annotated boxes
(\textbf{1,377} Pedestrian, \textbf{5,185} Vehicle, \textbf{344} Cyclist);
Day~2 contains \textbf{10,093} boxes
(\textbf{1,075} Pedestrian, \textbf{8,529} Vehicle, \textbf{489} Cyclist).
These ground-truth labels enable the downstream detection evaluation
described in Section~\ref{sec:detection}.

We will publicly release both test splits with annotations,
all model checkpoints, the fine-tuned SWIR detector, and training
code.  The complete training set will be released separately
following the completion of the required anonymisation process.

\vspace{-3mm}
\paragraph{Evaluation metrics.}
We report \textbf{FID}~\cite{heusel2017fid} and
\textbf{LPIPS}~\cite{zhang2018lpips} for image quality.
To measure the practical impact of detail preservation, we
complement these with a \textbf{downstream detection protocol}.
The rationale is that if synthetic SWIR is to serve as a proxy
for real SWIR in perception pipelines, a detector trained on
real SWIR should produce comparable results on both; a drop in
performance directly quantifies how much task-relevant structure
the translation has lost.  Concretely, we first fine-tune a
YOLO26l~\cite{jocher2023yolo} detector on real SWIR training images using the ground-truth
annotations, then run this detector on both real and generated SWIR
test images.  We evaluate the detections on generated images against
the ground-truth annotations using \textbf{mAP@50} and
\textbf{mAP@50-95}.  Lower scores indicate that the translation
has corrupted structural details needed for accurate perception.
We additionally report \textbf{size-stratified AP} using
COCO-style~\cite{lin2014coco} bins (small~$<32^2$,
medium~$32^2$--$96^2$, large~$\geq96^2$~px$^2$).

\vspace{-3mm}
\paragraph{Cross-dataset evaluation.}
To assess generalisation, we apply models trained on our dataset
\emph{zero-shot} to the public RASMD
benchmark~\cite{jin2025rasmd} (979 test pairs
captured in South Korea).  As the RASMD I2I translation benchmark does not provide detection
annotations, we report only FID and LPIPS for this evaluation.

\subsection{Baselines}
\label{sec:baselines}

We compare against:
(1)~\textbf{LDM}~\cite{rombach2022ldm}: latent diffusion
with the default source downsampling;
(2)~\textbf{LDiT}~\cite{peebles2023dit}: latent diffusion with a
Transformer denoiser, same conditioning as LDM;
(3)~\textbf{DDPM}~\cite{ho2020ddpm}: pixel-space diffusion
(memory-intensive but detail-preserving);
(4)~\textbf{LDM-LPL}~\cite{berrada2025lpl}: a standard LDM
trained with an additional latent perceptual loss to encourage
high-frequency detail;
(5)~\textbf{Pix2pix-turbo}~\cite{parmar2024i2iturbo} (abbreviated
P2P-turbo in tables): one-step latent diffusion adapted from a
frozen SD-Turbo~\cite{sauer2024sdturbo} backbone with
encoder--decoder skips.
We also report \textbf{Real SWIR} as a reference ceiling
(the detector applied to real SWIR images, evaluated against
ground-truth annotations).

All latent models use compression factor $f{=}8$.  For a fair
comparison, the first-stage autoencoder for every latent baseline
(LDM, LDiT, LDM-LPL) is a standard VQGAN trained from scratch on
the same SWIR training data as our SCAE; the only architectural
difference is the addition of $E^{\text{src}}$ and its skip
connections to~$D$. 
% Training uses four NVIDIA A100 GPUs (80\,GB),
% batch size 16, for 150 epochs per stage.
Inference uses DDIM~\cite{song2021ddim} with 200 steps.
Complete training hyperparameters are listed in the supplementary
material.

%%%%%%%%%%%%%%%%%%%%%%%%%%%%%%%%%%%%%%%%%%
\begin{table}[t]
\centering
\small
\setlength{\tabcolsep}{4pt}
\begin{tabular}{@{}llcccc@{}}
\toprule
& & \multicolumn{2}{c}{Day 1} & \multicolumn{2}{c}{Day 2 (shifted)} \\
\cmidrule(lr){3-4}\cmidrule(lr){5-6}
Method & Backbone & FID$\downarrow$ & LPIPS$\downarrow$ & FID$\downarrow$ & LPIPS$\downarrow$ \\
\midrule
LDM       & U-Net & 35.88 & 0.277$\pm$0.044 & 36.51 & 0.326$\pm$0.063 \\
DP-LDM    & U-Net & 19.01 & 0.158$\pm$0.041 & \underline{24.03} & \underline{0.216}$\pm$0.062 \\
\midrule
LDiT      & DiT   & 30.82 & 0.246$\pm$0.039 & 30.76 & 0.283$\pm$0.063 \\
DP-LDiT    & DiT   & \textbf{16.90} & \textbf{0.142}$\pm$0.042 & \textbf{21.57} & \textbf{0.192}$\pm$0.067 \\
\midrule
DDPM      & U-Net & \underline{18.34} & \underline{0.147}$\pm$0.040 & 29.19 & 0.287$\pm$0.082 \\
LDM-LPL      & U-Net & 35.14 & 0.276 $\pm$ 0.040 & 36.87 & 0.326 $\pm$ 0.065 \\
P2P-turbo & SD-Turbo & 21.56 & 0.184$\pm$0.032 & 25.11 & 0.223$\pm$0.061 \\
\bottomrule
\end{tabular}
\caption{Image quality.  Day~2 is recorded in a different area
and month (Section~\ref{sec:dataset}).
Best in \textbf{bold}, second-best \underline{underlined}.}
\label{tab:fid}
\vspace{-4mm}
\end{table}
%%%%%%%%%%%%%%%%%%%%%%%%%%%%%%%%%%%%%%%%%%

\subsection{Bottleneck Analysis}
\label{sec:bottleneck_exp}

Before comparing full translation pipelines, we isolate the two
sources of detail loss identified in Section~\ref{sec:bottleneck_method}.

\vspace{-4mm}
\paragraph{Autoencoder bottleneck.}
We encode real SWIR test images with the VQGAN encoder~$E$ and
immediately decode them with~$D$, bypassing the diffusion process
entirely.  Even in this best-case scenario (the decoder receives
a perfect latent code), the detector achieves only
0.702 mAP@50 (against ground-truth annotations) on the
reconstructions, compared to 0.856 on the original images.  This 18\% relative drop confirms that
compression alone destroys a measurable amount of task-relevant
detail.  Replacing the standard VQGAN with our SCAE
(Section~\ref{sec:scae}) nearly eliminates this loss, raising
reconstruction mAP@50 to 0.847, within 2\% of the real-SWIR
ceiling.
Both reconstruction ceilings are included in
Table~\ref{tab:detection}.

\vspace{-4mm}
\paragraph{Conditioning bottleneck.}
The full LDM pipeline (VQGAN + diffusion with naive resize
conditioning) achieves only 0.307 mAP@50, substantially worse
than even the VQGAN-only reconstruction (0.702).  This large gap
shows that the conditioning pathway is in fact the dominant source
of detail loss: the denoiser receives an impoverished version of
the source image and cannot recover the missing structure.

Taken together, these results confirm that both the autoencoder
and the conditioning pathway contribute to the detail loss
problem, and that addressing only one is insufficient.  The
following sections evaluate the full pipelines that combine
SCAE and LGE.

%%%%%%%%%%%%%%%%%%%%%%%%%%%%%%%%%%%%%%%%%%%%%%%%%%
\begin{figure}[!t]
\centering
\includegraphics[width=1.0\linewidth]{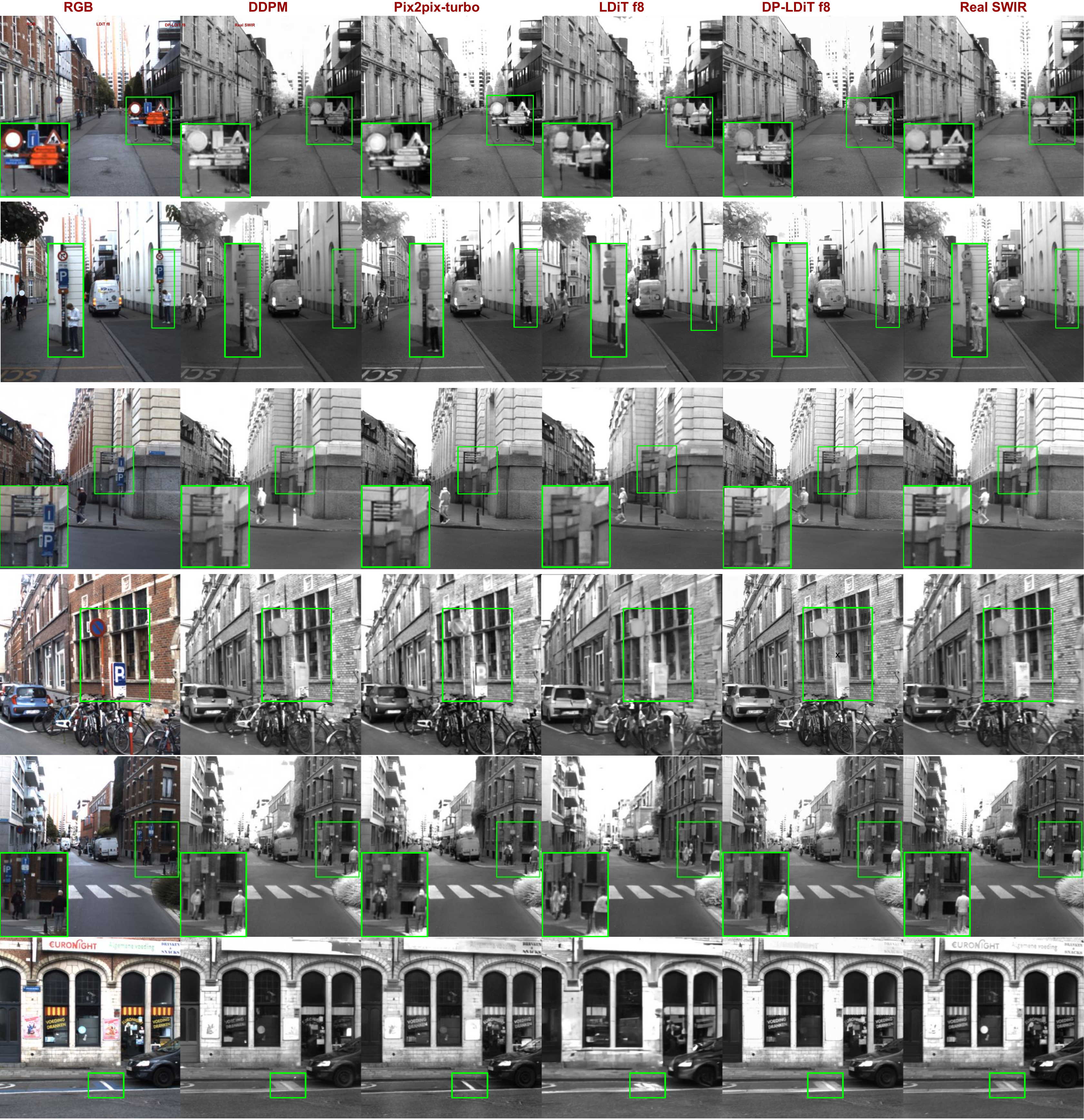}
% \caption{Qualitative comparison on Day~1.  Columns left to right:
% RGB input, DDPM, Pix2pix-turbo, LDiT~$f8$, DP-LDiT~$f8$ (ours),
% and real SWIR\@.  Green boxes highlight regions of interest.
% LDiT loses fine structure (signs, pedestrians); Pix2pix-turbo
% preserves structure but retains RGB-like appearance; DP-LDiT
% recovers details while faithfully rendering SWIR characteristics.
% Best viewed zoomed in.}
% \vspace{-5mm}
\caption{Qualitative comparison on Day~1. From left to right:
RGB, DDPM, Pix2pix-turbo, LDiT~$f$8, DP-LDiT~$f$8 (ours), and real SWIR.
Green boxes indicate regions of interest. DP-LDiT better preserves details
while matching SWIR appearance.}
\label{fig:qualitative}
\vspace{-4mm}
\end{figure}
% \vspace{-10cm}

%%%%%%%%%%%%%%%%%%%%%%%%%%%%%%%%%%%%%%%%%%%%%%%%%%
% \vspace{-9mm}
\subsection{Image Quality}
\label{sec:fid}

Table~\ref{tab:fid} reports FID and LPIPS on both test splits.
On Day~1, DP-LDiT achieves the best FID (\textbf{16.90}),
followed by DDPM (18.34) and DP-LDM (19.01).
Both DP variants substantially improve over their respective
baselines: DP-LDM reduces FID by 47\% relative to LDM
(19.01 vs.\ 35.88), and DP-LDiT reduces FID by 45\% relative
to LDiT (16.90 vs.\ 30.82).
On the shifted split (Day~2), all models degrade, but DP-LDiT
remains the strongest at 21.57 FID, while DDPM drops more
sharply to 29.19.  LPIPS trends mirror FID throughout.
Pix2pix-turbo achieves competitive FID (21.56) on Day~1 but,
as we show next, its generated images do not faithfully reflect
SWIR-specific appearance.

\paragraph{Qualitative comparison.}
Figure~\ref{fig:qualitative} presents example translations from
Day~1.  Because SWIR reflectance is governed by material
composition rather than visible-spectrum pigment, objects that
appear distinct in RGB can look similar in SWIR and vice versa.
Four observations illustrate this.

\emph{Traffic signs:} In real SWIR, painted markings prominent in
RGB (e.g.\ white text on blue parking signs) become faint or
invisible because the underlying paint shares similar SWIR
reflectance with the sign substrate.
DP-LDiT reproduces this correctly, whereas LDiT distorts sign
geometry into amorphous shapes and Pix2pix-turbo retains
RGB-like luminance contrast (e.g.\ markings remain clearly
visible), failing to capture the SWIR-specific appearance.

\emph{Pedestrians:} LDiT dissolves people into amorphous blobs,
making them unrecognisable as human figures.  DP-LDiT preserves
body contours, limb separation, and correctly renders the
SWIR-characteristic brightness that depends on clothing material
(e.g.\ cotton reflects more SWIR light than synthetic fabrics).
Pix2pix-turbo preserves body structure but produces unnatural
bright halo artefacts over dark clothing, a diffuse white fog
that does not match real SWIR appearance (rows~3, 5).

\emph{Ground markings:} Blue parking lines, which appear dark in
RGB but light grey in real SWIR, are correctly translated by all
diffusion methods including the LDiT baseline.  Pix2pix-turbo,
however, preserves the dark RGB appearance of these lines,
providing a clear example of modality leakage on a feature that
does not require fine spatial detail.

\emph{Vegetation:} Trees and bushes appear dark green in RGB but
bright in SWIR due to high leaf reflectance in this wavelength
range.  All diffusion baselines reproduce this shift correctly,
showing that coarse modality-specific appearance is relatively
easy to learn; the critical challenge lies in preserving fine
spatial structure such as pedestrians and sign text.

In summary, DP-LDiT recovers fine structure while maintaining
correct SWIR appearance across signs, people, ground markings, and vegetation.
Pix2pix-turbo preserves spatial structure well, owing to its
encoder--decoder skip connections, but consistently exhibits
appearance leakage, retaining RGB-like luminance patterns
rather than learning SWIR-specific reflectance properties.

%%%%%%%%%%%%%%%%%%%%%%%%%%%%%%%%%%%%%%%%%%
\begin{table}[t]
\centering
\resizebox{\linewidth}{!}{%
\begin{tabular}{@{}ll ccccc ccccc@{}}
\toprule
& & \multicolumn{5}{c}{Day 1} & \multicolumn{5}{c}{Day 2 (shifted)} \\
\cmidrule(lr){3-7}\cmidrule(lr){8-12}
Method & Type & mAP@50 & mAP@50-95 & AP\textsubscript{P} & AP\textsubscript{V} & AP\textsubscript{C}
                      & mAP@50 & mAP@50-95 & AP\textsubscript{P} & AP\textsubscript{V} & AP\textsubscript{C} \\
\midrule
Real SWIR  & ceiling  & 0.856 & 0.578 & 0.483 & 0.688 & 0.565 & 0.706 & 0.505 & 0.333 & 0.704 & 0.478 \\
VQGAN recon & AE ceiling & 0.702 & 0.440 & 0.289 & 0.596 & 0.436 & 0.527 & 0.352 & 0.146 & 0.587 & 0.323 \\
SCAE recon  & AE ceiling & 0.847 & 0.556 & 0.446 & 0.679 & 0.542 & 0.675 & 0.455 & 0.284 & 0.670 & 0.411 \\
\midrule
P2P-turbo  & SD-Turbo & \textbf{0.749} & \textbf{0.493} & \textbf{0.373} & \textbf{0.676} & \textbf{0.430} & \textbf{0.578} & \textbf{0.409} & \textbf{0.181} & \textbf{0.661} & \textbf{0.386} \\
DDPM       & Pixel    & 0.621 & 0.338 & \underline{0.203} & 0.560 & 0.252 & 0.513 & 0.290 & 0.105 & 0.500 & 0.264 \\
LDM        & Latent   & 0.307 & 0.155 & 0.040 & 0.358 & 0.069 & 0.241 & 0.130 & 0.013 & 0.312 & 0.065 \\
LDM-LPL    & Latent   & 0.332 & 0.170 & 0.052 & 0.378 & 0.081 & 0.256 & 0.139 & 0.015 & 0.320 & 0.082 \\
DP-LDM     & Ours     & 0.617 & 0.335 & 0.202 & \underline{0.567} & 0.238 & 0.548 & 0.309 & 0.130 & 0.537 & 0.261 \\
LDiT       & Latent   & 0.375 & 0.195 & 0.084 & 0.395 & 0.105 & 0.287 & 0.157 & 0.021 & 0.339 & 0.112 \\
DP-LDiT     & Ours     & \underline{0.631} & \underline{0.340} & 0.200 & 0.556 & \underline{0.265} & \underline{0.569} & \underline{0.320} & \underline{0.143} & \underline{0.541} & \underline{0.277} \\
\bottomrule
\end{tabular}%
}
\caption{Downstream detection.  A YOLO26l~\cite{jocher2023yolo} detector fine-tuned on real SWIR is applied to generated images; predictions are evaluated against ground-truth annotations.
VQGAN/SCAE recon rows show encode${\to}$decode reconstruction of real SWIR (no diffusion), establishing autoencoder ceilings.
Best generative in \textbf{bold}, second-best \underline{underlined}.}
\label{tab:detection}
\end{table}
% \vspace{-5mm}
%%%%%%%%%%%%%%%%%%%%%%%%%%%%%%%%%%%%%%%%%%

% \vspace{-3mm}
\subsection{Downstream Detection}
\label{sec:detection}

Table~\ref{tab:detection} presents detection metrics on both splits.
The results reveal a striking disconnect between FID and downstream
utility.

\vspace{-3mm}
\paragraph{Detail loss in latent diffusion cripples detection.}
On Day~1, LDM obtains only 0.307 mAP@50, far below the real-SWIR
ceiling of 0.856.  Training LDM with the auxiliary latent
perceptual loss (LDM-LPL~\cite{berrada2025lpl}) does not
substantially improve this (0.332 mAP@50, a mere $+8\%$ relative gain), confirming that
perceptual losses applied through noisy one-step approximations
are insufficient for preserving the structural details that matter
for downstream perception.
DP-LDM reaches 0.617, doubling detection
performance over the LDM baseline.  A similar improvement occurs for
LDiT (0.375 $\to$ 0.631 with DP-LDiT, a $1.7\times$ increase).  This confirms that the detail
loss in standard latent diffusion is not merely a perceptual
artefact but \emph{directly degrades downstream task performance}.
DDPM, operating in pixel space, achieves 0.621 mAP@50, comparable
to our latent-space DP variants, but at the cost of a batch size
of only 1 on an 80\,GB GPU (vs.\ 22 for DP-LDM).
The SCAE reconstruction ceiling (0.847,
Table~\ref{tab:detection}) shows that the autoencoder bottleneck
is nearly eliminated by our SCAE\@.  The remaining gap from
DP-LDiT (0.631) to that ceiling reflects the difficulty of
\emph{generating} faithful latent codes from noise, as opposed to
encoding real images.  This gap motivates both SCAE and LGE:
SCAE provides high-resolution source features that compensate for
imperfect latents at decode time, while LGE improves the latents
themselves.

%%%%%%%%%%%%%%%%%%%%%%%%%%%%%%%%%%%%%%%%
\begin{figure}[!t]
\centering
\includegraphics[width=0.9\linewidth]{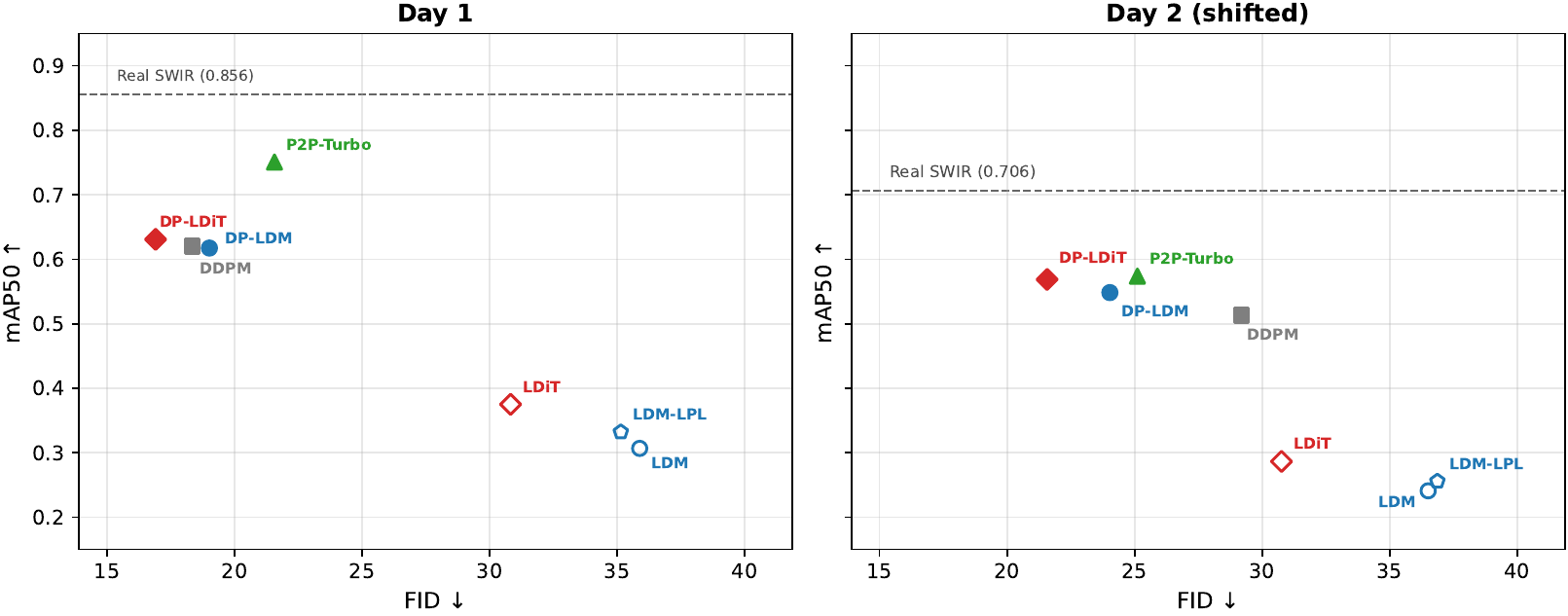}
\vspace{-3mm}
\caption{FID vs.\ mAP@50 on Day~1 (left) and Day~2 (right).
Arrows connect each baseline to its DP variant.  Dashed black
line: Real SWIR ceiling.  Pix2pix-turbo achieves high mAP
despite moderate FID, partly due to RGB-like outputs that the
detector handles well.  Our DP variants are the only methods
that move toward the ideal top-left corner (low FID, high mAP)
from both latent baselines.}
\label{fig:fid_vs_map}
\vspace{-4mm}
\end{figure}
%%%%%%%%%%%%%%%%%%%%%%%%%%%%%%%%%%%%%%%%
% \vspace{-5mm}
\vspace{-3mm}
\paragraph{FID does not predict detection performance.}
Figure~\ref{fig:fid_vs_map} visualises the FID--mAP relationship,
revealing poor correlation between the two metrics.  FID
computes distributional statistics over Inception features
pretrained on RGB ImageNet, which need not capture the
structures relevant for SWIR object detection.  Moreover,
these deep features are globally pooled, making them
sensitive to scene-level semantics but largely blind to
fine local structure such as small objects.  This
explains how LDM achieves an FID of 35.88 while its mAP
collapses to 0.307.

Conversely, Pix2pix-turbo attains the highest mAP (0.749) among
generative methods.  Two properties contribute:
(i)~end-to-end encoder--decoder skip connections jointly optimised
with the denoiser in a single inference step, allowing source
information to bypass both the latent bottleneck and the iterative
sampling process, unlike our SCAE, whose skip connections
operate only at the first-stage decoder and are frozen during
denoiser training. (ii)~Strong structural priors from its
SD-Turbo backbone, pretrained on billions of natural images,
which our models trained from scratch on ${\sim}$32K pairs cannot
match.  However, this frozen backbone also prevents learning
SWIR-specific appearance (Section~\ref{sec:fid}).  On Day~2, its
advantage narrows (0.578 vs.\ 0.569 for DP-LDiT), suggesting
that pretraining-driven structural priors generalise less well to
unseen conditions.

These results indicate that for perception-oriented translation,
FID should be treated as a necessary but insufficient condition:
it measures similarity between feature distributions extracted
by an ImageNet-trained classifier, but
task-level evaluation is needed to verify structural fidelity.
Our DP modifications are the only approach that simultaneously
improves \emph{both} axes.

\vspace{-3mm}
\subsection{Size-Stratified Analysis}
\label{sec:size}

Table~\ref{tab:size} breaks down detection AP by object size.
For large objects ($\geq 96^2$~px$^2$), most methods perform
similarly (AP $>0.91$), as coarse structure survives even lossy
compression.  The critical differences appear at small scale:
LDM achieves only 0.117 AP\textsubscript{small} while DP-LDM
reaches 0.392, a $3.4\times$ improvement.  DP-LDiT shows a similar
gain (0.144 $\to$ 0.392, $2.7\times$).  This confirms that our
modifications specifically recover the fine details that matter
for detecting small, distant objects such as pedestrians and
cyclists.

%%%%%%%%%%%%%%%%%%%%%%%%%%%%%%%%%%%%%%%%
% --- Ablation qualitative (suggested) ---
\begin{figure}[t]
\centering
\includegraphics[width=1.0\linewidth]{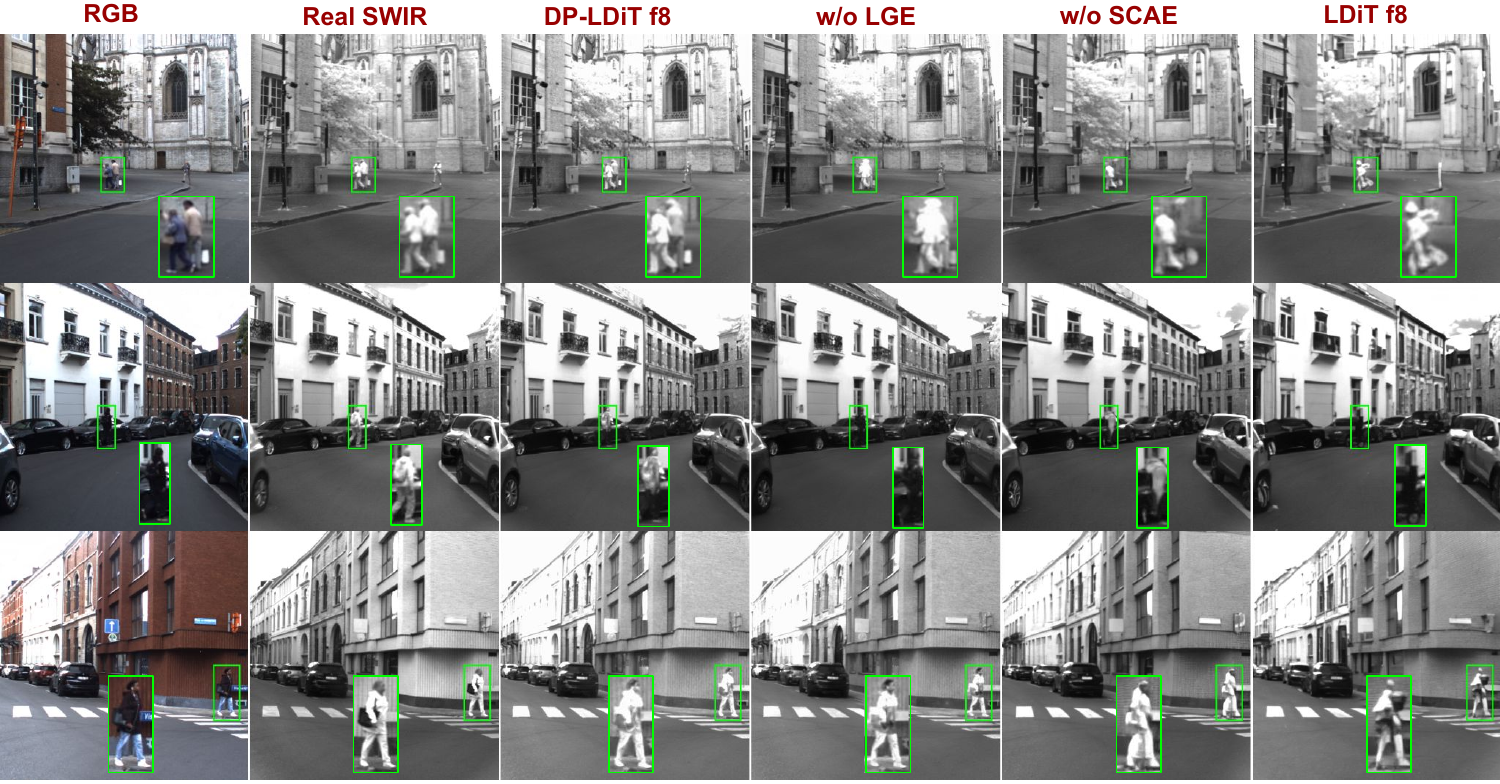}
\caption{Ablation examples on Day~1.  Columns: RGB input, real
SWIR, DP-LDiT (full model), w/o LGE (SCAE only), w/o SCAE
(LGE only), and LDiT baseline.  Green boxes highlight
pedestrians.  Best viewed zoomed in.}
\label{fig:ablation_qual}
\vspace{-4mm}
\end{figure}
%%%%%%%%%%%%%%%%%%%%%%%%%%%%%%%%%%%%%%%%

%%%%%%%%%%%%%%%%%%%%%%%%%%%%%%%%%%%%%%%%%%
\begin{wraptable}[10]{r}{0.40\textwidth}
\vspace{-1em}
\centering
\scriptsize
\setlength{\tabcolsep}{2.5pt}
\renewcommand{\arraystretch}{0.95}
\begin{tabular}{@{}lccc@{}}
\toprule
Method & AP\textsubscript{small}$\uparrow$ & AP\textsubscript{medium}$\uparrow$ & AP\textsubscript{large}$\uparrow$ \\
& ($n{=}2386$) & ($n{=}2797$) & ($n{=}883$) \\
\midrule
Real SWIR    & 0.739 & 0.901 & 0.940 \\
P2P-turbo    & \textbf{0.593} & 0.868 & 0.937 \\
DDPM         & 0.378 & 0.877 & \underline{0.939} \\
\midrule
LDM          & 0.117 & 0.474 & 0.749 \\
LDM-LPL      & 0.154 & 0.487 & 0.752 \\
DP-LDM       & \underline{0.392} & \underline{0.879} & 0.935 \\
\midrule
LDiT         & 0.144 & 0.586 & 0.914 \\
DP-LDiT      & \underline{0.392} & \textbf{0.885} & \textbf{0.945} \\
\bottomrule
\end{tabular}
\caption{Size-stratified AP@50 on Day~1.}
\label{tab:size}
% \vspace{-20cm}
\end{wraptable}
%%%%%%%%%%%%%%%%%%%%%%%%%%%%%%%%%%%%%%%%%%

\subsection{Ablation Study}
\label{sec:ablation}

Table~\ref{tab:ablation} isolates the contributions of LGE and
SCAE on both backbones.  Starting from the LDM baseline
(FID 35.88, mAP 0.307):
adding LGE alone reduces FID to 24.48 and raises mAP to 0.470
(+53\% relative);
adding SCAE alone gives FID 22.56 and mAP 0.594 (+94\%);
combining both yields FID 19.01 and mAP 0.617 (+101\%).
The pattern is consistent for the DiT backbone, where the full
DP-LDiT achieves 16.90 FID and 0.631 mAP (+68\% over LDiT baseline).

SCAE contributes more than LGE in isolation, even though the
bottleneck analysis (Section~\ref{sec:bottleneck_exp}) shows
the conditioning pathway causes the larger mAP drop
(0.702$\to$0.307) compared to autoencoder compression
(0.856$\to$0.702).  This is because SCAE's skip connections
serve a dual role: they not only raise the autoencoder
reconstruction ceiling (from 0.702 to 0.847) but also compensate
for imperfect latent codes from the denoiser by providing
high-resolution source features directly to the decoder.  LGE
addresses the conditioning bottleneck by improving the latent
codes the denoiser produces.  The two components are
complementary because they operate at different stages: LGE
improves what latent code the denoiser produces, while SCAE
improves how the decoder renders it into pixels.

Figure~\ref{fig:ablation_qual} illustrates these differences
visually.  The LDiT baseline (rightmost column) produces severely
degraded output in which pedestrians are barely recognisable.
\emph{Without LGE (SCAE only).}
Removing LGE recovers spatial structure through the skip
connections, yet the model struggles with semantic parsing:
nearby pedestrians merge into indistinguishable blobs (row~1),
dark-clothed figures disappear against dark backgrounds (row~2),
and clothing boundaries are incorrectly segmented (row~3).
\emph{Without SCAE (LGE only).}
Layout and object placement are preserved, but fine spatial
detail is degraded: edges are softer and small features
lose definition.
These complementary roles explain why combining both
consistently yields the best results: LGE helps the denoiser
\emph{understand} the scene while SCAE helps the decoder
\emph{render} fine detail.

\begin{table}[t]
\centering
\small
\setlength{\tabcolsep}{3.5pt}
\begin{tabular}{@{}llcccccc@{}}
\toprule
Method & Backbone & LGE & SCAE & FID$\downarrow$ & LPIPS$\downarrow$ & mAP50$\uparrow$ & mAP50-95$\uparrow$ \\
\midrule
Baseline     & U-Net & -- & -- & 35.88 & 0.277 & 0.307 & 0.155 \\
+\,LGE       & U-Net & \checkmark & -- & 24.48 & 0.195 & 0.470\,{\scriptsize(+53\%)} & 0.250\,{\scriptsize(+61\%)} \\
+\,SCAE      & U-Net & -- & \checkmark & 22.56 & 0.183 & 0.594\,{\scriptsize(+94\%)} & 0.319\,{\scriptsize(+106\%)} \\
DP-LDM       & U-Net & \checkmark & \checkmark & \textbf{19.01} & \textbf{0.158} & \textbf{0.617}\,{\scriptsize(+101\%)} & \textbf{0.335}\,{\scriptsize(+116\%)} \\
\midrule
Baseline     & DiT & -- & -- & 30.82 & 0.246 & 0.375 & 0.195 \\
+\,LGE       & DiT & \checkmark & -- & 22.91 & 0.178 & 0.506\,{\scriptsize(+35\%)} & 0.270\,{\scriptsize(+38\%)} \\
+\,SCAE      & DiT & -- & \checkmark & 20.14 & 0.169 & 0.615\,{\scriptsize(+64\%)} & 0.334\,{\scriptsize(+71\%)} \\
DP-LDiT       & DiT & \checkmark & \checkmark & \textbf{16.90} & \textbf{0.142} & \textbf{0.631}\,{\scriptsize(+68\%)} & \textbf{0.340}\,{\scriptsize(+74\%)} \\
\bottomrule
\end{tabular}
\caption{Ablation on Day~1.  Percentages show relative
mAP@50 improvement over the respective baseline.}
\label{tab:ablation}
% \vspace{-30mm}
\end{table}

%%%%%%%%%%%%%%%%%%%%%%%%%%%%%%%%%%%%%%%%%%%%%%%%%%%%%%%%%
\begin{table}[t]
\centering
\small
\begin{tabular}{@{}llcccc@{}}
\toprule
& & \multicolumn{2}{c}{Our dataset} & \multicolumn{2}{c}{RASMD (zero-shot)} \\
\cmidrule(lr){3-4}\cmidrule(lr){5-6}
Method & Backbone & FID$\downarrow$ & LPIPS$\downarrow$ & FID$\downarrow$ & LPIPS$\downarrow$ \\
\midrule
LDM    & U-Net & 35.88 & 0.277 & 78.13 & 0.498 \\
DP-LDM & U-Net & \underline{19.01} & \underline{0.158} & \underline{71.71} & \underline{0.468} \\
LDiT   & DiT   & 30.82 & 0.246 & 72.39 & 0.496 \\
DP-LDiT & DiT   & \textbf{16.90} & \textbf{0.142} & \textbf{68.67} & \textbf{0.455} \\
\bottomrule
\end{tabular}
\caption{Cross-dataset generalisation.  Models trained on our
dataset, evaluated zero-shot on RASMD\@.
Best in \textbf{bold}, second-best \underline{underlined}.}
\label{tab:crossdataset}
\vspace{-4mm}
\end{table}
%%%%%%%%%%%%%%%%%%%%%%%%%%%%%%%%%%%%%%%%%%%%%%%%%%%%%%%%%
%%%%%%%%%%%%%%%%%%%%%%%%%%%%%%%%%%%%%%%%
\begin{figure}[!t]
\centering
\includegraphics[width=1.0\linewidth]{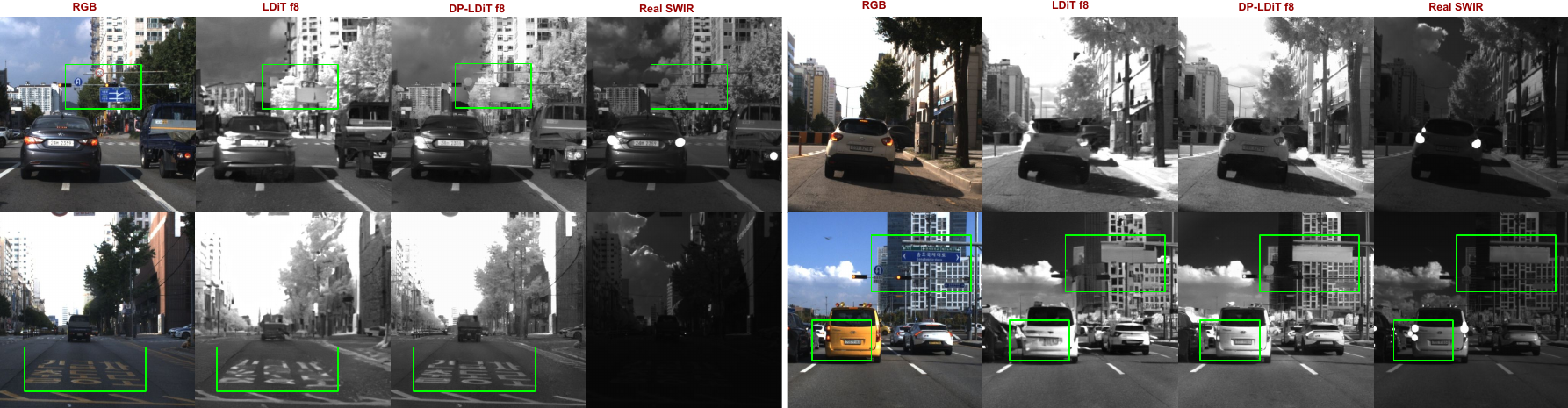}
\vspace{-7mm}
\caption{Zero-shot cross-dataset results on RASMD\@.  Models
trained on our dataset are applied without fine-tuning.
Despite the overall darker SWIR appearance (due to different
camera hardware), DP-LDiT preserves fine details and correctly
translates SWIR-specific properties such as sign reflectance
and vegetation brightness.}
\label{fig:crossdataset_qual}
\vspace{-4mm}
\end{figure}
%%%%%%%%%%%%%%%%%%%%%%%%%%%%%%%%%%%%%%%%

% \vspace{-3mm}
\subsection{Cross-Dataset Generalisation}
\label{sec:crossdataset}

Table~\ref{tab:crossdataset} reports zero-shot FID and LPIPS on
the RASMD benchmark using models trained exclusively on our dataset.
Despite the domain gap (different country, camera hardware, and
driving conditions), DP-LDiT achieves 68.67 FID, improving over the
LDiT baseline (72.39) and LDM (78.13).  The consistent ranking
across datasets confirms that SCAE and LGE address fundamental
bottlenecks in the latent diffusion pipeline rather than overfitting
to training-data-specific patterns.

The higher absolute FID on RASMD (69 vs.\ 17) likely reflects
the combined effect of different SWIR camera hardware, geographic
setting, and weather conditions.  Nevertheless, Figure~\ref{fig:crossdataset_qual} shows
that DP-LDiT correctly generalises SWIR properties: sign
reflectance, vegetation brightness, and fine structural details
are well preserved, while LDiT loses them.  This confirms that
our modifications capture SWIR imaging physics rather than
memorising dataset-specific appearance.  Additional zero-shot
examples are shown in the supplementary material.

%%%%%%%%%%%%%%%%%%%%%%%%%%%%%%%%%%%%%%%%%%%%%%%%%%%%%%
\begin{wraptable}[11]{r}{0.48\linewidth}
\vspace{-1.2em}
\centering
\scriptsize
\setlength{\tabcolsep}{2.5pt}
\renewcommand{\arraystretch}{0.95}
\begin{tabular}{@{}lrrrr@{}}
\toprule
Model & Backbone & Params (M) & Size (GB) & Time (s)$\downarrow$ \\
\midrule
DDPM       & U-Net & 34.0  & 0.13 & 19.36 \\
\midrule
LDM        & U-Net & 309.2 & 1.15 & 3.62 \\
DP-LDM     & U-Net & 358.4 & 1.34 & 3.64 \\
\cmidrule{2-5}
\multicolumn{2}{@{}l}{Overhead (U-Net)} & +49.2 & +0.19 & +0.02 \\
\midrule
LDiT       & DiT   & 525.6 & 1.96 & 3.93 \\
DP-LDiT    & DiT   & 564.9 & 2.11 & 4.01 \\
\cmidrule{2-5}
\multicolumn{2}{@{}l}{Overhead (DiT)} & +39.3 & +0.15 & +0.08 \\
\bottomrule
\end{tabular}
\caption{Model complexity (FP32) and inference time per image
(DDIM, 200 steps, single A100).}
\label{tab:efficiency}
\vspace{-0.8em}
\end{wraptable}
%%%%%%%%%%%%%%%%%%%%%%%%%%%%%%%%%%%%%%%%%%%%%%%%%%%%%%
\subsection{Efficiency}
\label{sec:memory}

Table~\ref{tab:efficiency} summarises model sizes and inference
times.  The parameter overhead of our modifications is modest:
+49.2\,M (+0.19\,GB) for DP-LDM and +39.3\,M (+0.15\,GB) for
DP-LDiT, coming primarily from $E^{\text{src}}$ in the SCAE and
the lightweight LGE ($<0.1$\,M).  DDPM is the smallest
(34.0\,M, 0.13\,GB) as it needs no autoencoder, but pixel-space
diffusion at $512 \times 512$ limits training to batch size~1
on an 80\,GB GPU, whereas our latent-space DP variants support
substantially larger batches on the same hardware.

Inference cost follows a similar pattern.  DP-LDiT takes 4.01\,s
per image, only 0.08\,s (2.0\%) slower than the LDiT baseline
(3.93\,s), yet $4.8\times$ faster than DDPM (19.36\,s) while
achieving comparable detection performance (0.631 vs.\ 0.621
mAP@50).

%--------------------------------------------------------------
\section{Discussion}
\label{sec:discussion}
%--------------------------------------------------------------

\paragraph{Why does detail loss hurt detection so much?}
The size-stratified analysis (Table~\ref{tab:size}) reveals that
standard LDMs primarily fail on small objects, whose features
span only a few latent tokens and are therefore most vulnerable
to compression artefacts.  Large objects (AP $>0.91$ for all
methods) are relatively robust because their features span
multiple tokens and survive lossy encoding.  This asymmetry
explains the FID--mAP disconnect: FID's globally pooled deep
features capture scene-level statistics but are largely blind
to fine local structure, remaining moderate even when
small-object detail is destroyed.
For perception-oriented translation, FID should be treated as
necessary but insufficient; complementary task-level metrics
are essential.  We also note that our protocol evaluates
detections against ground-truth annotations: two methods with
similar mAP may succeed on different objects.  Directly
comparing detection sets on real and translated images
(without ground truth as intermediary) would provide a
finer-grained view of translation quality.

\vspace{-3mm}
\paragraph{SCAE vs.\ LGE.}
SCAE provides the larger individual boost.  The
bottleneck analysis (Section~\ref{sec:bottleneck_exp}) shows that
SCAE nearly eliminates the autoencoder bottleneck (reconstruction
mAP rises from 0.702 to 0.847), but its benefit extends beyond
reconstruction: the skip connections also compensate for imperfect
latent codes at decode time.  LGE addresses the conditioning
bottleneck, which causes the larger raw mAP drop
(0.702$\to$0.307) but is partially mitigated by SCAE's bypass.  Notably, the auxiliary latent perceptual
loss of LDM-LPL~\cite{berrada2025lpl} yields only a marginal
improvement over LDM (0.332 vs.\ 0.307 mAP@50), reinforcing that
the bottleneck is architectural, not a matter of training
objectives.  The two components are complementary: LGE improves
\emph{what} the denoiser generates while SCAE improves \emph{how}
the decoder renders it.

\vspace{-3mm}
\paragraph{Does SCAE simply copy RGB features?}
One might ask whether the skip connections in SCAE merely pass
RGB appearance through to the output, bypassing the modality
translation.  This is not the case: $E^{\text{src}}$ is trained
end-to-end with $D$ under a VQGAN reconstruction loss on SWIR
images, so the skip features are optimised for SWIR
reconstruction, not RGB passthrough.  The qualitative results
confirm this: white markings on blue traffic signs, which are
prominent in RGB, correctly disappear in our outputs (consistent
with real SWIR), and vegetation appears bright following SWIR
reflectance rather than the darker green of RGB\@.  A model that
simply copies source features would retain these RGB-specific
visual cues.

\vspace{-3mm}
\paragraph{Backbone agnosticism.}
The consistent improvements across U-Net and DiT backbones
(Tables~\ref{tab:fid}--\ref{tab:ablation}) confirm that LGE and
SCAE address a bottleneck in the autoencoder and conditioning
pipeline, not in the denoiser architecture.  This suggests
applicability to future backbone designs without modification.

\vspace{-3mm}
\paragraph{Limitations.}
Our method requires paired, spatially registered training data
and is evaluated on RGB-to-SWIR translation only, though SWIR
is a particularly suitable testbed as it captures reflected light
at RGB-comparable resolution, making the detail bottleneck most
apparent.  The approach learns the RGB-to-SWIR mapping implicitly
and cannot generalise to materials underrepresented in training,
as illustrated by complex commercial signage that none of the
evaluated methods reproduce faithfully
(Figure~\ref{fig:qualitative}, row~6); incorporating material-level
conditioning from a foundation segmentation model could improve
robustness.  Finally, we evaluate translated images only as a
stand-in for real SWIR in a detection benchmark; whether they are
equally useful as training data augmentation, where distributional
coverage may matter more than per-image fidelity, remains open.

%--------------------------------------------------------------
\section{Conclusion}
\label{sec:conclusion}
%--------------------------------------------------------------

We identified two distinct detail bottlenecks in latent diffusion
I2I translation: the autoencoder compression and the conditioning
pathway.  To address them, we proposed two light\-weight and
backbone-agnostic modifications: a Source-Conditioned Auto\-encoder
(SCAE) and a Learnable Guidance Encoder (LGE).  Applied to
RGB-to-SWIR translation on our paired driving dataset, DP-LDM
and DP-LDiT improve downstream detection mAP by up to
$2\times$ compared to their base models, with up to $3.4\times$
gains on small objects, while achieving state-of-the-art
FID\@.  The improvements generalise across denoiser backbones
and transfer zero-shot to the RASMD
benchmark~\cite{jin2025rasmd}.

Beyond the architectural contributions, our evaluation reveals that
FID and detection performance are poorly correlated in this setting,
highlighting the need for task-level evaluation of I2I translation
in safety-critical applications.  Detail preservation is not just
a visual quality concern; it has direct, measurable impact on
downstream perception.

\bibliography{refs}

\end{document}